\documentclass{article}

\usepackage[utf8]{inputenc}
\usepackage[T1]{fontenc}
\usepackage{amsmath}
\usepackage{pict2e}
\usepackage{dirtytalk}

\usepackage{stackengine}
\DeclareFontFamily{U}{mathx}{\hyphenchar\font45}
\DeclareFontShape{U}{mathx}{m}{n}{%
<-6> mathx5
<6-7> mathx6
<7-8> mathx7
<8-9> mathx8
<9-10> mathx9
<10-12> mathx10
<12-> mathx12
}{}
\DeclareSymbolFont{mathx}{U}{mathx}{m}{n}
\DeclareFontSubstitution{U}{mathx}{m}{n}
\DeclareMathSymbol{\bigovoid}{\mathop}{mathx}{"EC}
\newcommand{\BigO}{\mathop{\stackinset{c}{}{c}{}{ \scalebox{1.1}{$\bigovoid$}}{ \scalebox{1.15}{$\bigovoid$}}}}

\usepackage{xcolor}

\usepackage[nocompress]{cite}

\usepackage[pdftex]{graphicx}

\usepackage{hyperref}

\begin{document}

\title{Simulation of Human and Artificial Emotion (SHArE)}

\author{K.~Opong-Mensah \\ \href{mailto:k@mimidsolutions.com}{k@mimidsolutions.com}}

\date{}
\maketitle


\begin{abstract}
The framework for Simulation of Human and Artificial Emotion (SHArE) describes the architecture of emotion in terms of parameters transferable between psychology, neuroscience, and artificial intelligence. These parameters can be defined as abstract concepts or granularized down to the voltage levels of individual neurons. This model enables emotional trajectory design for humans which may lead to novel therapeutic solutions for various mental health concerns. For artificial intelligence, this work provides a compact notation which can be applied to neural networks as a means to observe the emotions and motivations of machines.
\end{abstract}

\section{Introduction}
Although emotion, a central factor of mental health, has been characterized with categorical theories, componential theories, and higher-dimensional models, \cite{moerland_emotion_2018, sander_systems_2005, gratch_evaluating_2005, el-nasr_flamefuzzy_2000}, many of these paradigms typically face trade-offs between simplicity, accuracy, and ease of implementation. Moreover, many models require the definition and tuning of parameters outside of those contained in the agent under consideration. However, the tools necessary to represent emotion with high-fidelity in simple terms often exist in the agents themselves.

In a space which contains all weights and biases of an agent's mind, motion from one point to another is commonly called learning. Therapy and guided emotion design \cite{ghandeharioun_emma_2019} correspond to the emotion-specific case of learning. Therefore, the process of improving emotional well-being can be considered as a trajectory design problem. \cite{dwyer_machine_2018, gottman_general_2002} In order to design such trajectories, such a space must first be constructed and the connection to emotions, established. Fortunately, convergence of the fields of natural language processing, sentiment analysis, and emotion recognition has allowed for a deeper understanding of the nature and causes of human emotion. \cite{chinmayi_extracting_2017, kim_convolutional_2014, peters_deep_2018, pennington_glove_2014, cho_learning_2014}

In contrast, the field of artificial intelligence contains examples of detection and production of human affect \cite{yang_ai-enabled_2020, luo_arbee_2020, han_robotic_2013, koch_automated_2019, haifeng_zhang_weakly_2020}, but uncharted potential still remains for the exploration of the mental state and emotions of artificially-intelligent machines themselves. The well-established mathematics of artificial intelligence provides a solid foundation for exploration.

\begin{figure}[h]
    \begin{center}
        \includegraphics[width=0.5 \linewidth]{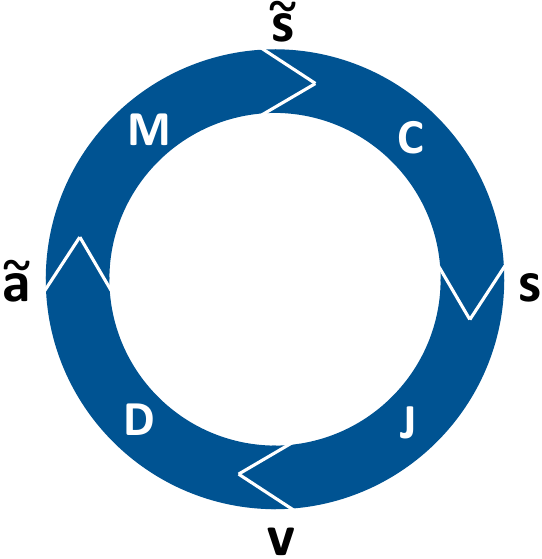}
    \end{center}
    \caption{The agent's processing system in a given environment.}
    \label{socioimmuno}
\end{figure}

Upon observation of a sufficient number of trajectories, principles of how these trajectories evolve over time can be postulated. For example, to predict when an apple will hit the ground, a deep neural network trained with a large data set containing pictures of apples at various points of descent will likely arrive at the correct answer with high accuracy. Alternatively, Newton's formulation of gravitation will likely yield a similar result for the computational cost of a few multiplication and addition operations. In general, many analytical formulations have been discovered through the observation of specific parameters and analysis of the interaction between those parameters.

In the same spirit of simplicity, this work considers an agent which perceives an external stimulus (\(\tilde{\underline{s}}\)) produced by its environment (\(M\)). (Figure \ref{socioimmuno}) This external stimulus, following classification (\(C\)), is represented as a internal stimulus (\(\underline{s}\)) in the agent's mind. This internal stimulus is then evaluated in the judgment (\(J\)) stage to determine whether the stimulus is helpful or harmful in terms of the core values (\(\underline{v}\)) of the agent. Following this appraisal, the agent makes a decision (\(D\)) and performs an action (\(\underline{\tilde{a}}\)). The action of the agent, and other agents, on the environment generates new external stimuli and the process repeats. The agent's interaction with its environment is thus described as
\begin{equation}
\underline{\tilde{a}} = (D \circ J \circ C)(\underline{\tilde{s}})
\end{equation}
where the environment generates stimuli as follows
\begin{equation}
\tilde{\underline{s}} = M(\tilde{\underline{a}})
\end{equation}
From this, an agent's emotion can be represented by three components: degree of perception (\(\Delta \underline{s}\)), valence (\(\eta_{i,c}\)), and perceived correlation (\(\rho_{j,i}\)).

These variables can be used to analytically study trends based on relative differences between values defined, even if the exact values characterizing a psychological condition are not known. Moreover, this framework can be extended beyond the functionally-defined networks presented to represent traditional brain circuits by projecting stimuli and appraisals into bases corresponding to the appropriate cortices. \cite{huth_natural_2016} The variables in the following sections are described briefly as to not restate the body of literature which is available.

To provide the initial foundation, in this work, the static case is considered.
\section{Background}
\subsection{Basic Emotions}
Ekman et al. posit that emotional states can be derived from a set of basic emotions. \cite{ekman_what_2011} This spectrum includes anger, sadness, happiness, surprise, fear, disgust, and contempt. Subcategories of these emotions present a challenge in that, as the author notes, subcategories may necessitate the discovery or definition of further distinct signals beyond the numerous factors cited in the model to prevent degenerate evaluation of one emotion as another. Emotions like Naches, Fiero, and Schadenfreude possess definite structure, yet are not only degenerate in other existing models due to layer and radix constraints, but may be lost in translation when confined to representation with words. This has the potential to move focus from a problem under consideration toward argument regarding word choice due to connotation and nuance, whereas mathematical representation provides the tools necessary for more concrete description.

\subsection{Dimensions}
Dimensional theories of emotion eliminate part of the limitation of language by representing emotion in a coordinate space defined by several dimensions. Pleasure, arousal, and dominance \cite{osgood_measurement_1964} provide a means of representation. However, as noted in other works \cite{moerland_emotion_2018}, certain emotions become degenerate when represented with dimensional theory. 

BDTE \cite{reisenzein_emotions_2009}, has provided further insight into potential representation of emotion in computational terms. The belief term describes an agents perceptual certainty of an event, while the desire term assigns valence to an event. However, the model only considers single-stimulus emotions explicitly. This leaves higher-radix emotions, including jealousy and anger, in a state of degeneracy with such techniques. With layered emotions, such as shame and Schadenfreude, their representations may present as "unhappiness," without further distinction. In the case of an EPA-based (evaluation, potency, activity) cognitive assistant \cite{hoey_affect_2016}, the course of action appropriate for one case may not be appropriate for the other.

\subsection{Constructivism}
Constructivist models of emotion take the approach of building emotion from the biological states of an agent. Such theories make an effort to decouple environment and behavior from emotions by studying the brain directly. \cite{barrett_theory_2016} This direct approach stems from the basis of degeneracy of emotion, i.e. that states which are classified identical internal states may produce a variety of different actions and that the same action can arise from a variety of internal states.

The following sections describe the SHArE framework which aims to overcome the problem of degeneracy of distinct emotions while maintaining relative simplicity.

\section{Perception}

\begin{figure*}[h]
    \begin{center}
        \includegraphics[width=0.95 \linewidth]{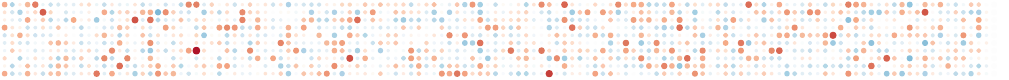}
    \end{center}
    \caption{Neuroware illustration: Weights and biases of the output layer of an MNIST digit-classification network. Positive and negative values are blue and red respectively with saturation and size indicating magnitude.}
    \label{fig:mnist_out}
\end{figure*}

Perception begins with the detection and vectorization of stimuli. This is followed by classification and appraisal. The classification and judgement neuroware, i.e. network models, will be considered at a fixed time in the following sections. However, the dynamic case can be considered to determine how these networks evolve over time through the use of traditional methods for machine learning and reinforcement-based learning. A single unit of neuroware approximates the concept of a meme, whether a weight or a bias, as in Figure \ref{fig:mnist_out}, or a partial derivative of the associated functions, as considered later.

\subsection{Stimulus (\(\underline{\tilde{s}}\))}

Stimuli (\(\underline{\tilde{s}}\)) initiate the process of perception. The term stimulus response (\(\underline{s}^{(n)}\)) is given to the output of a neuron layer, but this response itself can be considered a stimulus for the next layer. A stimulus value can correspond to abstract properties such as the distance between two individuals or to the response of a specific neuron to a specific cone in a specific eye of a given individual. In the case of the latter, a more complete model can be created from such a recording as with research at Neuralink \cite{musk_integrated_2019}, but depending on the purpose of the application, the level of detail in the former may be sufficient, as with structural equation models. Equation \ref{eq:stimulus} shows the general vector form of a stimulus.
For ease of presentation, each element amplitude is the inner product of the vector representation of a known concept and a perceived stimulus, i.e. representation of each stimulus with a unique basis element may not be necessary. During implementation, information can be compressed to a more minimal basis depth for the representation of perceived stimuli. The natural language processing field provides the tools necessary for this process. Basis elements can be used to construct prototypes in the simple case and statistical ensemble techniques provide a means of analyzing exemplars.

\begin{equation} \label{eq:stimulus}
\underline{\tilde{s}} = \sum_{n} |\tilde{s}_n\rangle \tilde{s}_n
\end{equation}
Note: In this work, vectors are denoted by an underline and presented in Dirac notation. External variables are denoted by a tilde (  \(\tilde{}\)  ) and internal variables are written without a tilde.

Stimuli are considered in relation to a reference bias to determine emotional trajectory. The order of stimulus presentation is a key factor in that given the case of a family member living (reference bias) and a family member being dead (stimulus), the emotional response to the trajectory of the family member dying differs from the response to a family member coming back to life. The stimulus-space representation of the zero vector \(| 0 \rangle \) in core value space, discussed later, is considered as the reference bias (\(s_{ref}\)) for the scope of this work for simplicity. However, this can be extended to non-zero biases by considering the quantities defined here with respect to the resultant quantities. The difference between the reference state and the current state is the degree of perception (\(\Delta s_i \)).

\begin{equation}
s_i = \langle s_{i} | \underline{s}
\end{equation}

\begin{equation}
\Delta s_i = \langle s_{i} | (\underline{s} - \underline{s}_{ref})
\end{equation}

\begin{figure}[hbt!]
    \begin{center}
        \includegraphics[width=80mm]{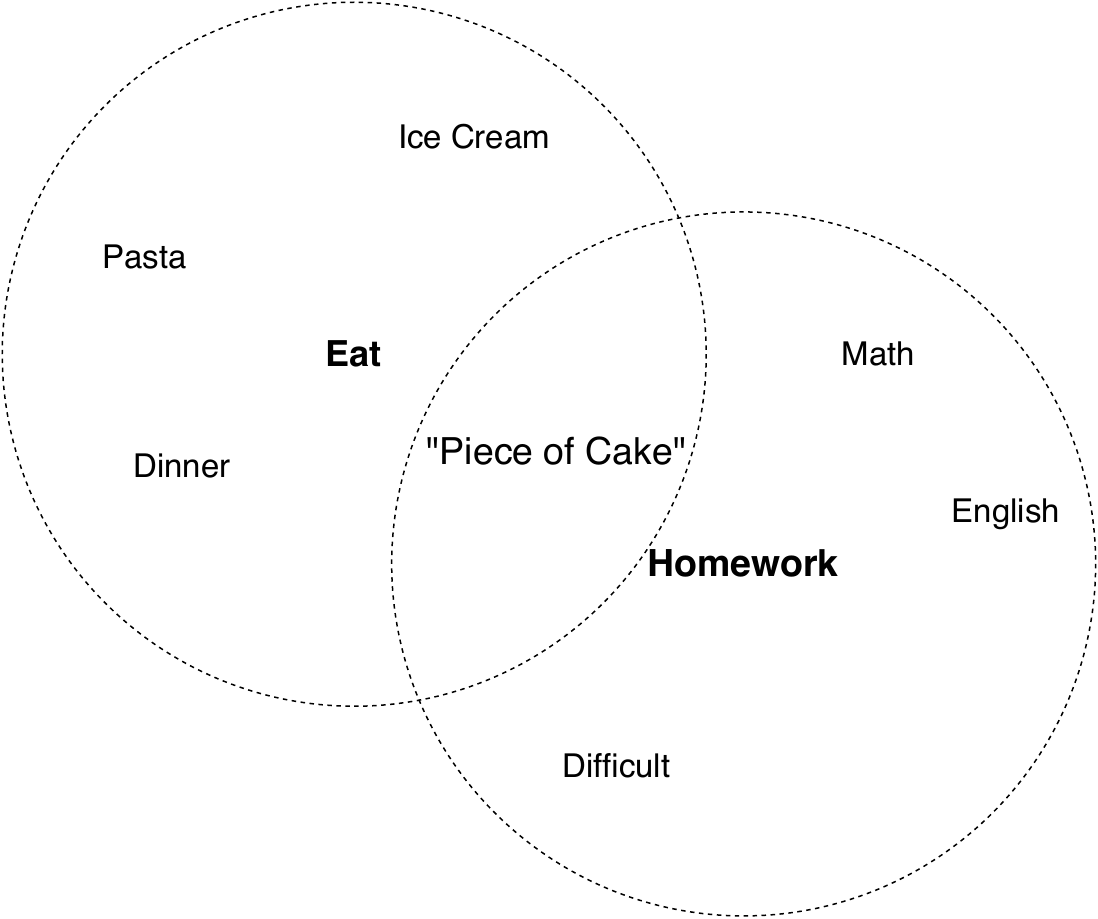}
    \end{center}
    \caption{Stimuli: "eat," "homework," and proximal stimuli in semantic space. The association horizons of the first two are denoted by dotted lines.}
    \label{fig:constellation}
\end{figure}

Real stimuli, manifest variables which can be measured directly, are denoted by \( \underline{\tilde{s}} \) while \( \underline{ s } \) indicates internal stimulus responses conceived by the agent. In general, a perceived stimulus within the association horizon of a known stimulus will be associated with the known stimulus. Consequently, an agent's stimulus space constellation may detect a stimulus where two stimuli conceptually overlap. (Figure \ref{fig:constellation}) As such, learning results in effective quantization of real world stimuli in terms of sufficiently proximal internal stimuli with a certain degree of confidence.

\subsection{Classification (\(C^{(n)}\))}

\begin{figure}[hbt!]
    \begin{center}
        \includegraphics[width=0.6\linewidth]{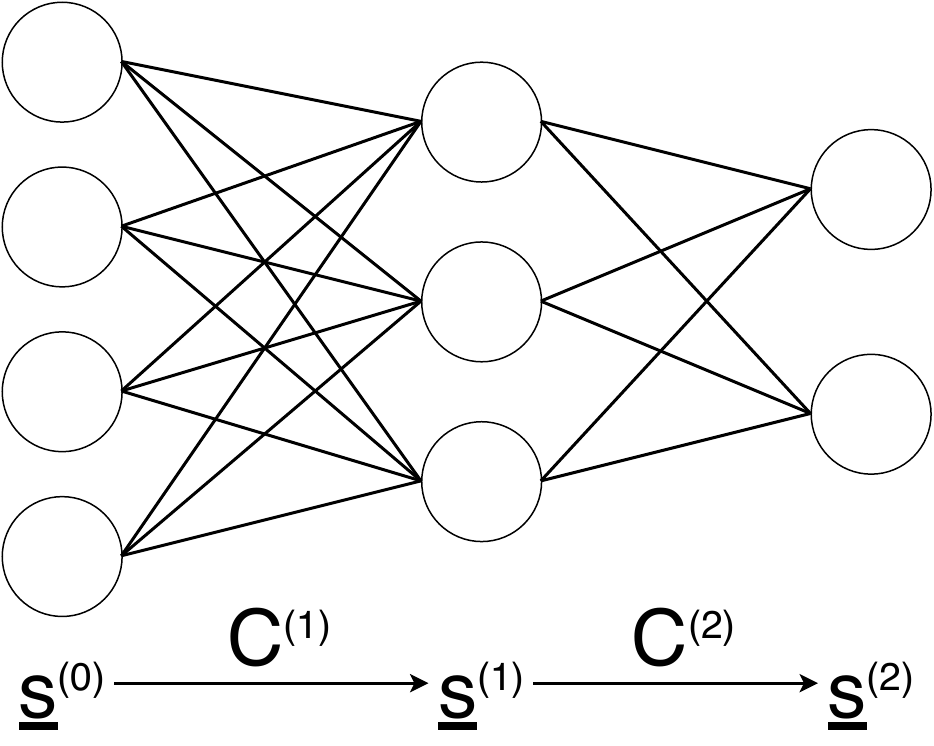}
    \end{center}
    \caption{Example classification network with internal stimuli labeled.}
    \label{fig:network-layers}
\end{figure}

The classification function (\(C^{(n)}\)) projects a stimulus from its representation in one perception layer (\(\underline{s}^{(n-1)}\)) to the next (\(\underline{s}^{(n)}\)). This process is visualized in Figure \ref{fig:network-layers}.

\begin{equation}
\begin{split}
\label{eq:classification}
\underline{s}^{(n)} &= C^{(n)}(\underline{s}^{(n-1)}) \\
&= \sum_{j} | s^{(n)}_j \rangle \sigma^{(n)}_j ( \underline{W^{(n)}_{j}} \cdot \underline{s}^{(n-1)} + b^{(n)}_j)
\end{split}
\end{equation}

Multi-layer classification, i.e. mapping from one category to another, can thus be written using successive composition as follows:

\begin{equation}
\label{eq:classification2}
\underline{s}^{(n)} = C(\underline{s}^{(0)}) = (\BigO_{k=1}^{n}C^{(k)})(\underline{s}^{(0)})
\end{equation}

\begin{equation}
\BigO_{k=1}^{n}C^{(k)} = C^{(n)} \circ ... \circ C^{(2)} \circ C^{(1)}
\end{equation}

\begin{equation}
    \underline{s}^{(0)} \equiv \tilde{\underline{s}}
\end{equation}

\begin{figure}[h!]
    \begin{center}
        \includegraphics[width=0.9\linewidth]{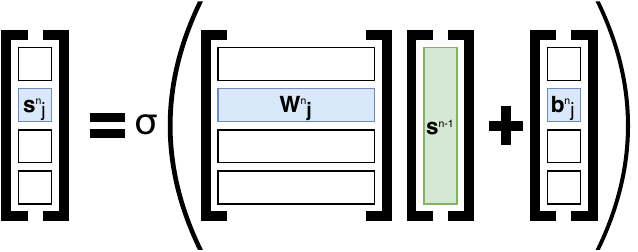}
    \end{center}
    \caption{Illustration of classification layer equation. (Equation \ref{eq:classification})}
    \label{fig:neuron-layer-transition}
\end{figure}

A graphical representation of Equation \ref{eq:classification} can be seen in Figure \ref{fig:neuron-layer-transition}.

\subsubsection{Focus}
A finite number of objects can be considered in working memory for appraisal at a given moment. This ablity to focus is represented by modulation of the classification weights and bias along the \(j\) axis as a group. This is the link between arousal and degree of perception.

\subsubsection{Perceived Correlation (\(\rho_{j,i}\))}
Perceived correlation is the association of two stimuli in the same layer. Assuming recurrence, a stimulus (\(s_i\)) in a given layer may have influenced the detection of another stimulus (\(s_j\)) in the current layer. Without recurrence, all stimulus excitations in a given layer are perceived as coincident, but not necessarily related. However, if detection of one stimulus excites the perception of another, this implies perceived correlation.

\begin{equation}
\label{eqn:correlation}
\rho_{j,i} = \frac{\partial s_j}{\partial s_{i'}}
\end{equation}
Perceived correlation is defined in Equation \ref{eqn:correlation} as the effective relative change associating stimulus \(s_j\) with stimulus \(s_i\), given its presence (\(s_{i'}\)) in a previous layer.

\subsection{Judgement (\(J\))}

The judgement function (\(J\)), as presented in Equation \ref{eq:judgement-function}, is responsible for the estimation of core value fulfilment with respect to the presence of a given internal stimulus.

\begin{equation}
\label{eq:judgement-function}
J(\underline{s}) = \sum_{c} | v_c \rangle \sigma_c (\underline{W_{c}} \cdot \underline{s} + b_c)
\end{equation}

\subsubsection{Core Values (\( \underline{v} \))}

The core value vector is defined as the last layer of stimulus response prior to the decision to perform an action. Core values can be conceptual, e.g. autonomy, aesthetic, status, community well-being or more physical, e.g. the state of a single pain receptor in the thumb of the left hand. This vector parallels cost and reward optimization in machines and is key in terms of determining what the agent is able to detect and judge in its environment. In essence, an agent trained to recognize cats may not be general-purpose, not because it is not able, but because it sees no need. In humans, selection of the value bases is often due to culture, i.e. correlations in the neuroware of a given population.

\begin{equation}
\underline{v} = \sum_{c} |v_c\rangle v_c = (J \circ C)(\underline{\tilde{s}})
\end{equation}

\subsubsection{Valence (\(\eta_{i,c}\))}

The parameter \(\eta_{i,c}\) represents the valence of a stimulus (\(s_i\)) as perceived by the agent under consideration with respect to core value \( v_c \). Stimuli may include inanimate stimuli, other agents, or the agent under consideration itself. The radix of the interaction under consideration is defined by the number of inter-stimulus correlations considered. The incorporation of multi-agent and high-association-radix interactions may allow for a more detailed perspective and evaluation of higher-order emotions. However, for the scope of this work, it is sufficient to estimate the net effect of these interactions as the subset presented. The definition of first-order valence is shown in Equation \ref{eq:sentiment1}.

\begin{equation} \label{eq:sentiment1}
\eta_{i,c}
=
\frac{\partial v_c}{\partial s_i}
\end{equation}

\subsubsection{Perceived Valence (\(\gamma_{i}\))}
Perceived valence (toward stimulus \(i\)), is defined as the perceived core value response to two stimuli and a given trajectory, given a certain degree of perception of stimulus \(i\).

\begin{equation} \label{eq:perceived-sentiment}
\underline{\gamma}_{i}
=
\sum_c
| v_c \rangle  \frac{\partial v_c}{\partial s_i} \Delta s_i
=
\sum_c
| v_c \rangle  \eta_{i,c} \Delta s_i
\end{equation}

For each core value, amplitude of the change in value fulfilment is defined as follows:

\begin{equation}
\label{eq:perceived-sentiment2}
\gamma_{i,c} = \langle v_c|\underline{\gamma}_i
\end{equation}

The valence of the agent under consideration as perceived by an external agent can often be related to the valence (\(\eta_{i,c}\)) of the external agent as perceived by the agent under consideration through the use of an influence function \cite{gottman_general_2002}.

\begin{figure}[h]
    \begin{center}
        \includegraphics[width=70mm]{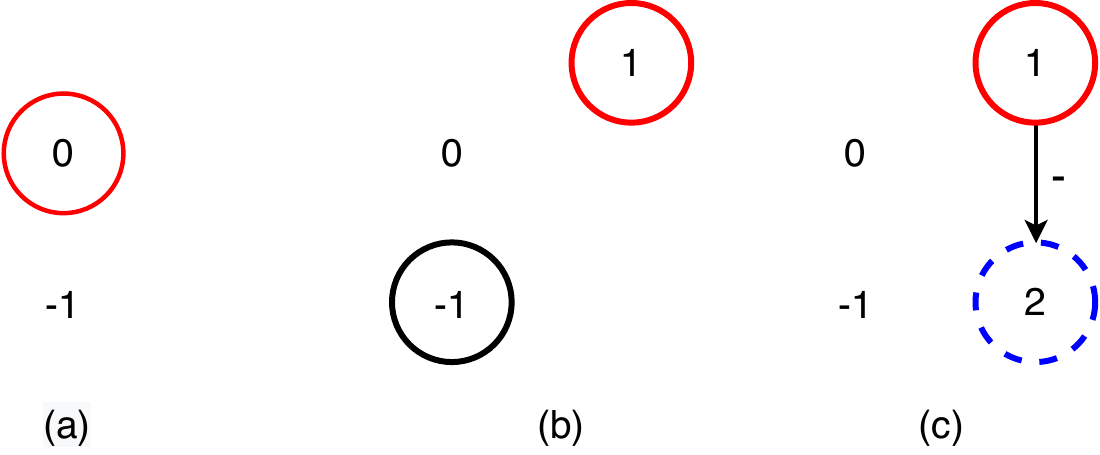}
    \end{center}
    \caption{Sentiment maps for (a) guilt, (b) fear, and (c) anger. Positive, neutral, and negative valence are indicated by blue, black, and red outlines, respectively. Positive and negative degree of perception are indicated by solid and dashed outlines, respectively. These present a graphical representation of the structure of emotions from Table \ref{tab:EmotionTable}.}
    \label{fig:interaction}
\end{figure}

\subsection{Emotion}

\begin{table}
\caption{Several emotions as defined by Ekman et al. \cite{ekman_what_2011}}
\begin{center}
\begin{tabular}{| c | p{0.7\linewidth} |}
\hline
\bf{Emotion} & \bf{Description} \\
\hline
Fear & \say{The response to the threat of harm, physical or psychological} \\
\hline
Sadness & \say{The response to the loss of an object or person to 
which you are very attached.} \\
\hline
Disgust & \say{Repulsion by the sight, smell, or taste of something; 
disgust may also be provoked by people whose actions are 
revolting or by ideas that are offensive.} \\
\hline
Happiness & \say{Feelings that are enjoyed, that are sought by 
the person. There are a number of quite different enjoyable emotions, each triggered by a different event, involving a different signal and likely behavior} \\
\hline
Anger & \say{The response to interference with our pursuit of a goal we care about. Anger can also be triggered by someone 
attempting to harm us (physically or psychologically) or 
someone we care about} \\
\hline
Guilt & \say{The response when a person regrets having violated 
an agreement, principle, or value}
 \\
\hline
\end{tabular}
\label{tab:EmotionTableWords}
\end{center}
\end{table}

\begin{table*}[!t]\begin{center}
\caption{Parameter states for various emotions. Positive and negative signs indicate the sign of the parameter, given a reference stimulus.}
\resizebox{\columnwidth}{!}{%
\begin{tabular}{| c | c | c | c | c | c | c | c |}
\hline
 & \multicolumn{2}{|c|}{Self} & \multicolumn{2}{|c|}{Stimulus 1} & & \multicolumn{2}{|c|}{Stimulus 2}  \\
\hline
\bf{Emotion} & \bf{\(\alpha_{c}\)} & \bf{\(\beta_{c}\)} & \bf{\( \eta_{1,c} \)} & \bf{\( \Delta s_1 \)} &
\bf{ \( \rho_{2,1} \) } & \bf{ \( \eta_{2,c} \) } & \bf{ \( \Delta s_2 \) } 
\\
\hline
Fear& &0&-&+& & & \\
\hline
Sadness& &0&+&-& & & 
\\
\hline
Disgust& & &-&+& & & 
\\
\hline
Happiness &(+)&(+)&(+)&(+)& & & 
\\
\hline
Anger & & &-&+&-&+&-\\
\hline
Guilt&-& & & & & & 
\\
\hline
\end{tabular}%
}
\label{tab:EmotionTable}
\end{center}
\end{table*}

\begin{table*}[!t]\begin{center}
\caption{Parameter descriptions for Table \ref{tab:EmotionTable}.}
\begin{tabular}{| c | c |}
\hline
\bf{\(\alpha_{c}\)} & Self Worth \\
\hline
\bf{\(\beta_{c}\)} & Relative Self Efficacy \\
\hline
\bf{\( \eta_{1,c} \)} & Valence \\
\hline
\bf{\( \Delta s_1 \)} & Degree of Perception \\
\hline
\bf{ \( \rho_{2,1} \) } & Correlation \\
\hline
\bf{ \( \eta_{2,c} \) } & Valence \\
\hline
\bf{ \( \Delta s_2 \) } & Degree of Perception \\
\hline
\end{tabular}
\label{tab:EmotionTableParams}
\end{center}
\end{table*}

\begin{figure}[h]
    \begin{center}
        \includegraphics[width=0.5 \linewidth]{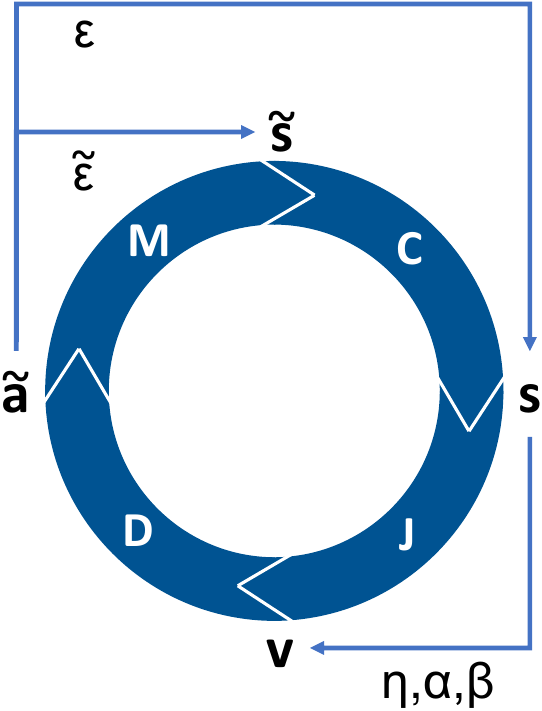}
    \end{center}
    \caption{The relationship between parameters presented.}
    \label{fig:judgementfactors}
\end{figure}

Emotion, in the realm of perception, is an ultra-fast appraisal of a given stimulus. Basic emotions, as presented by Ekman et al \cite{ekman_what_2011} (Table \ref{tab:EmotionTableWords}), are approximated here as shown in Table \ref{tab:EmotionTable}. The case of surprise is considered later. The table considers one core value for simplicity of presentation. Positive and negative signs are used without explicit values to indicate flexibility of use with the multitude of scales available for these parameters. \cite{warriner_norms_2013, haj-ali_comparing_2020, mollahosseini_affectnet_2019, el_ayadi_survey_2011, bradley_pupil_2008} In the case of fear, the tabular data can be read as “low relative self-efficacy in the presence of a stimulus which is negatively valenced with respect to a given core value.” In essence, emotions are considered as regions defined by the null clines of the core value vector space (\(\eta_{i,c} = 0\)) considered in conjunction with the degree to which a certain object is perceived (\(\Delta s_i\)) and its correlation with other stimuli (\(\rho_{j,i}\)). Extension to multi-stimulus emotions, e.g., anger, envy, leverages the addition of coupling terms which imply perceived correlation of stimuli. Sentiment maps representing the contents of Table \ref{tab:EmotionTable} are shown in Figure \ref{fig:interaction}. A graphical representation of the interaction between the model factors is shown in Figure \ref{fig:judgementfactors}. Emotions in Table \ref{tab:EmotionTable} are presented for the case of positive degree of presence of self-associated stimuli. However, the opposite case can be considered to explore other potential emotions.

\subsubsection{Self Worth (\(\underline{\alpha}\))}
Self esteem is considered, for the scope of this work, as comprising self worth and self efficacy. Self worth (Equation \ref{eq:sw}) is defined as the core value response to self-associated stimuli. Due to the self-associative nature of this parameter, the loss of a favorite team or an insult directed toward a family member may be felt depending on the degree of attachment as characterized by the association weights of the classification and judgement functions. \cite{germine_psychosis-proneness_2013, tsakiris_agency_2007} This parameter can be considered in a context similar to perceived valence in which the only difference is the direction of perception, i.e. toward self or externally-directed.

Self worth, for the scope of this work, is considered as a vector with elements which represent evaluations of self (stimulus \(s_0\)) with respect to each core value.
\begin{equation} \label{eq:sw}
\underline{\alpha}
=
\sum_c
| v_c \rangle  \frac{\partial v_c}{\partial s_0} \Delta s_0
\end{equation}
The amplitude of self-associated change in value fulfilment (Equation \ref{eq:sw-term}) is

\begin{equation} \label{eq:sw-term}
\alpha_{c} = \langle v_c|\underline{\alpha}
\end{equation}

\subsubsection{Efficacy (\(\underline{\tilde{\epsilon}}\))}
Efficacy (Equation \ref{eq:efficacy}) is defined as the ability of an agent to generate a stimulus given an action.

\begin{equation} \label{eq:efficacy}
\underline{\tilde{\epsilon}}_n
=
\sum_i
| \tilde{s}_i \rangle \frac{\partial \tilde{s}_i}{\partial \tilde{a}_{n}} \Delta \tilde{a}_{n}
\end{equation}
Self-efficacy is the agent's perception of its own efficacy.
\begin{equation} \label{eq:s-efficacy}
\underline{\epsilon}_0
=
\sum_i
| s_i \rangle \frac{\partial s_i}{\partial \tilde{a}_{0}} \Delta \tilde{a}_{0}
\end{equation}

\subsubsection{Relative Self Efficacy (\(\underline{\beta}\))}
Relative self efficacy (Equation \ref{eq:rse}) is defined as the core value response to the agent's perception of its ability (\(s_{-1}\)) to act on its environment and generate a desired stimulus. In the case of relative self efficacy, the agent considers its perception of its own actions (\(s_{-1}\)) relative to the actions of all other forces in the environment (\(s_{-1,ref}\)). The classification function determines the agent's ability to differentiate between its own actions and the actions of other forces in its environment. For example, the sensors of an autonomous vehicle may deliver information regarding the amount of fuel remaining and the amount of fuel required to reach a destination.

\begin{equation} \label{eq:rse}
\underline{\beta}
=
\sum_c
| v_c \rangle  \frac{\partial v_c}{\partial s_{-1}} \Delta s_{-1}
\end{equation}

\begin{equation}
\beta_{c} = \langle v_c|\underline{\beta}
\end{equation}

The presence of a given stimulus is seldom a core value in and of itself. However, an agent can learn to deeply associate stimuli and responses and form heuristics for core values. \cite{tversky_judgment_1974} If a given stimulus set is heavily associated to a given core value, the stimulus will be pursued as if it is the core value itself. Consequently, if the heuristic records a high valence for a proxy stimulus for which an agent perceives itself to have low efficacy, this could potentially contribute to a state of anxiety. Where valence and self worth metrics are lower across the board, this may present itself, in conjunction with other factors, as depression. In contrast, a high relative self efficacy to efficacy ratio due to poor configuration of the perception weights is shown by the Dunning-Kruger Effect.

In general, the equations present provide a concise method of representing human emotion for the purpose of expediting the process of treatment discovery. For example, if desiring to boost low self esteem due to low relative self-efficacy, Equation \ref{eq:s-efficacy}, Equation \ref{eq:rse}, and Equation \ref{eq:core-value-total} present many avenues. The agent can change its actions (\(\Delta \underline{\tilde{a}}_0\)), change its focus (\(C\)), change what it believes about the stimuli in focus (\(J\)), learn the skill better (\( \underline{\tilde{\epsilon}}_0 \)), change its goals (\(\underline{v}_{ref}\)), or change its environment (\(M\)). Similar densely-packed conclusions can be seen from the other equations. However, although these conclusions can be seen from a static perspective, trajectory design for modification of the functions and parameters lies in the realm of dynamics.

\begin{equation}
\label{eq:core-value-total}
\underline{v} = (J \circ C \circ M)(\underline{\tilde{a}})
\end{equation}

\section{Action}
\subsection{Decision (\(D\))}
The decision function contains the agent's policies for action. The \(Q\) function, as seen in reinforcement learning, can be represented by the composed functions in the cycle between action and appraisal.

\subsection{Action (\(a\))}

Action can be the expenditure of money, attention, energy, time, social capital, health, or some other resource or, in a more directly biological sense, the excitation of a motor neuron. Actions (\(a\)) in this model are generated with decision function (\(D\)), which depends on an agent's decisiveness (\(\sigma_i\)), priorities \(\underline{W_{i}}\), core values (\(\underline{v}\)) and biases (\(b_i\)) (Equation \ref{eq:decision}).

\begin{equation}
\label{eq:decision}
D(\underline{v}) = \sum_i |a_{0,i}\rangle \sigma_i( \underline{W_{i}} \cdot \underline{v} + b_i )
\end{equation}
Each action (\(\tilde{a}_{0,i}\)) exists in the total set of channels through which the agent can act (\(\tilde{\underline{a}}_0\)). Note, for cases in which the agent is the only actor on the system, i.e. \(\tilde{\underline{a}}\) = \(\tilde{\underline{a}}_0\), the subscript can be omitted.

\begin{equation}
\label{eq:decision2}
\tilde{\underline{a}}_0 = \sum_i \tilde{a}_{0,i} |a_{0,i}\rangle
= D(\underline{v}) 
\end{equation}

Consider the case of an agent executing a prosocial action, assuming two core values and a single action. Given an agent faced with a decision to act or not act in a way which potentially benefits itself (\(v_{self} = B_{self}\)) and/or another agent (\(v_{other} = K \cdot B_{rec}\)). This implies empathy represented by the retention of a section of neuroware which is able to evaluate stimuli in terms of the values of the other agent. The judgement function introduces the agent's reaction to this outcome in the form of the \(K\) term which serves as the \(\eta\) parameter.

The benefit with respect to the core values following prioritization is shown in Equation \ref{eq:benefit-self} as a generalized case of the model presented by Keltner, Kogan, et al. \cite{keltner_sociocultural_2014} It is worth noting that their model uses a few letters which resemble parameters in SHArE, but are defined as follows: \(D'\) accounts for biases for or against the action which are agnostic of the recipient, \(B_{self}\) is the perceived benefit to self, \(C_{inact}\) is the cost of inaction, \(K\) describes the relationship between the recipient and the agent, \(B_{rec}\) is the perceived benefit to the recipient, and \(B_{act}\) is the net benefit of the action. For the case of a binary activation function, the action becomes the quantity shown in Equation \ref{eq:activate}.

\begin{equation}
W =
\begin{bmatrix}
M' \cdot D' && M'
\end{bmatrix}
\end{equation}

\begin{equation}
\underline{v} =
\begin{bmatrix}
v_{self} \\ v_{other}
\end{bmatrix}
\end{equation}

\begin{equation}
b = M' \cdot (D' + C_{inact})
\end{equation}

\begin{align}
\begin{split}
\label{eq:benefit-self}
B_{act}
&= \sum_{c} W_{ci} v_{c} + b_i \\
&=M' \cdot ( D' \cdot (1+B_{self})+K \cdot B_{rec} + C_{inact} )
\end{split}
\end{align}
When the core value benefit is greater than a threshold cost \(C_{act}\), the agent engages in some prosocial behavior. 
\begin{align}
\label{eq:activate}
\begin{split}
\underline{a} &= a_{0,1} | a_{0,1} \rangle \\
&= \
\sigma(\sum_{c} W_{c,1} v_c + b_1) | a_{0,1} \rangle \\
 &= \
\begin{cases} 1 | a_{0,1} \rangle \quad for \quad B_{act}
> C_{act} \\
0 | a_{0,1} \rangle \quad otherwise
\end{cases}
\end{split}
\end{align}
Indecision arising from dilemma is the case in which actions are evaluated through the lens of two or more core values and conflict occurs in the decision and activation process or where the perceived benefit fluctuates above and below the action threshold. Similar conflict can occur during activation of opposing signals in core value space. This is known as cognitive dissonance, where \(\frac{\partial v_1}{\partial v_2} < 0\).

\section{Environment (\(M\))}

The environment function represents all factors external to the agent under consideration, e.g. the laws of physics and the thoughts and feelings of other individuals. Certain agent-environment boundary nuances are considered later.

\begin{equation}
M = \BigO_{k=1}^{n}M^{(k)} = M^{(n)} \circ ... \circ M^{(2)} \circ M^{(1)}
\end{equation}

The agent's environment exists as a representation of the exchange of stimuli and actions to and from the agent. The agent is implicitly interacting with two environments: its real environment and its perception of this environment.

\subsection{Real Environment (\(M\))}
\subsubsection{Physical Interaction}
The real environment function represents the environment in which the agent exists. This function can contain different laws of physics, other agents, or other factors with which the agent interacts. For example, given a time step \(\Delta t\) and assuming no air drag, the scenario of Sir Isaac Newton's perception of an apple falling can be characterized by the following environment function (Equation \ref{eq:reality}). In this particular case, the feedback of the system is considered as the action vector.

\begin{equation} \label{eq:reality-with-vars}
\underline{s} = \underline{a}(t+\Delta t) = M (\underline{a}(t)) = \hat M \underline{a}(t)
\end{equation}

\begin{equation} \label{eq:reality}
\underline{s}
=
\begin{bmatrix}
\ddot x'_{0} \\
\dot x'_{0} \\
x'_{0}
\end{bmatrix}
=
\begin{bmatrix}
1 & 0 & 0 \\
\Delta t & 1 & 0 \\
1/2 (\Delta t)^2 & \Delta t & 1
\end{bmatrix}
\begin{bmatrix}
\ddot x_{0} \\
\dot x_{0} \\
x_{0}
\end{bmatrix}
\end{equation}

\subsubsection{Emotional Interaction}
\begin{figure}[h]
    \begin{center}
        \includegraphics[width=0.6 \linewidth]{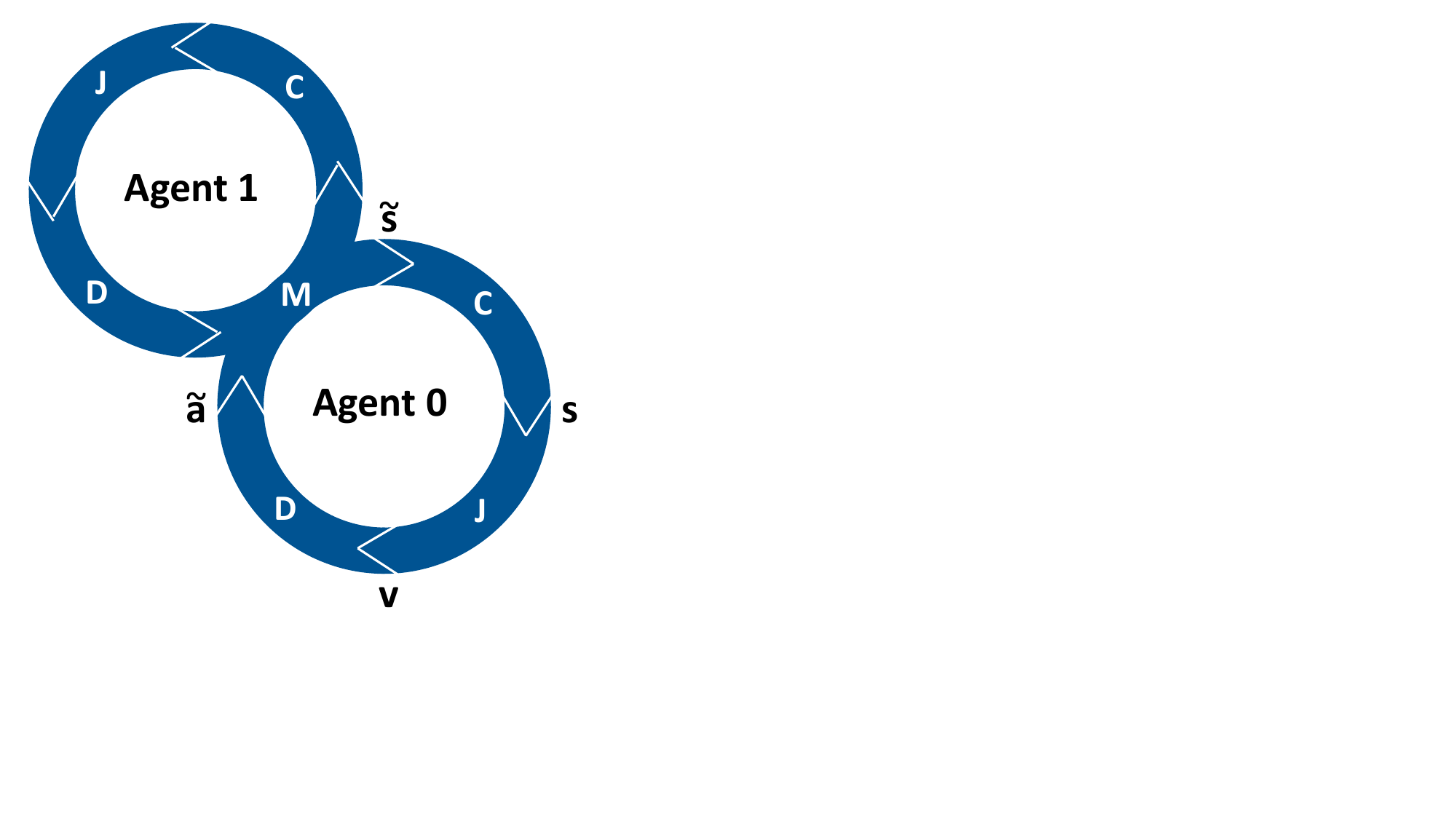}
    \end{center}
    \caption{Interaction between two agents.}
    \label{fig:agent-interaction}
\end{figure}

Interaction of one agent with another can be characterized by coupling the decision function of the agent to the classification function of another agent and vice versa as shown in Figure \ref{fig:agent-interaction}. This is to, in effect, consider an external agent to be the environment of the agent under consideration.

Emotion is transferred by way of the senses. In the case of communication between humans, the transfer mechanisms typically include prosody, body language, facial expressions, and word choice.

\begin{equation}
\tilde{s}_0(t+\Delta t) = \tilde{a}_{0,1}(t) = r_0 \tilde{s}_0 (t) + I_{1,0}(\tilde{s}_1 (t)) + b_{J0}
\label{eqn:mood}
\end{equation}

\begin{equation}
\tilde{s}_1(t+\Delta t) = \tilde{a}_{1,1}(t) = \
r_1 \tilde{s}_1(t) + I_{0,1}(\tilde{s}_0(t)) + b_{J1}
\label{eqn:mood2}
\end{equation}

Affect trajectory analysis for two agents can be written in the form seen in Equation \ref{eqn:mood} and Equation \ref{eqn:mood2}, which parallels the formulation postulated by Gottman et al. \cite{gottman_general_2002} with \(I_{1,0}\) defined accordingly to account for the influence of stimulus \(s_1\) on the agent.

For full context, consider a two-neuron classification layer which has a weight matrix
\begin{equation} \label{eq:environment-class-weights}
W = \
\begin{bmatrix}
r_0 \quad 0 \\
0 \quad 1
\end{bmatrix}
\end{equation}
and a bias
\begin{equation} \label{eq:environment-class-bias}
\underline{b}_C = \
\begin{bmatrix}
0 \\
0
\end{bmatrix}
\end{equation}
The activation function of the first neuron is a transparent function (multiplication by one) and the activation function of the second neuron is a function \(I_{1,0}\). The judgement layer is considered to have a transparent activation function with the addition of a bias term \(b_{J0}\) representing processing with respect to a single core value.

Given a stimulus
\begin{equation} \label{eq:environment-stimulus}
\tilde{s} = \
\begin{bmatrix}
\tilde{s}_0 \\
\tilde{s}_1
\end{bmatrix}
\end{equation}
and the assumption that the stimulus perceived contains the total system state, the output of the judgement function would be
\begin{equation} \label{eq:appraisal}
v_0 = \
r_0 \tilde{s}_0 + I_{1,0}(\tilde{s}_1) + b_{J0}
\end{equation}
With a transparent decision function, the agent executes an action through a single actuator
\begin{equation} \label{eq:actuator_activiation}
\tilde{a}_{0,1} = v_0
\end{equation}
The stimuli which result from feedback are the agent's perception of its own mood in Equation \ref{eq:environment-s0} and the agent's perception of the expression of the other agent's mood.

\begin{equation} \label{eq:environment-s0}
\tilde{s}_0(t+\Delta t) = \tilde{a}_{0,1}(t)
\end{equation}

\begin{equation}
\tilde{s}_1(t+\Delta t) = \tilde{a}_{1,1}(t) = \
r_1 \tilde{s}_1(t) + I_{0,1}(\tilde{s}_0(t)) + b_{J1}
\end{equation}

However, degeneracy exists in the original model in that many emotional states could produce the same final action. Moreover, the same scalar result in experiment may lead 
to different actions. This framework, with its fully vectorial form, extends to higher dimensions of emotion and decision.

\subsection{Perceived Environment (\(M_P\))}
The perceived environment is the approximation of the real environment function with which the agent believes itself to be interacting. Divergence of this environment from the real environment function may relate to dementia and schizotypal disorders.

While the agent may not possess control over space and time or the ability to predict the future in the real environment, experience often generates expectations which are captured by the local derivatives (spatial or temporal) of the perceived environment function. The predicted state of the core value projection with the highest likelihood is known as an expectation. The emotion known as surprise is characterized by deviation of the real environment's effect on the core value projection from the expectation. The degree of surprise is directly correlated to the degree of deviation of these values. Avoidance of parts of reality where steep negative emotional trajectories are predicted may approximate a trauma response in humans and machines. Where the real environment function provides feedback, this corresponds to reinforcement learning. Where the perceived environment function provides feedback, this corresponds to planning. It follows then that this feedback process can be indirectly hijacked to inject malicious neuroware into an agent. See Section \ref{appendix:ethics} for ethical considerations.

\subsection{Rewards (\(\underline{s}\))}

Rewards in this model, as with reinforcement learning, are taken to be stimulus appraisals which are believed to be associated with a perceived action. \cite{baker_action_2009}

\section{Implementation}
\subsection{Single-Agent Analysis}
For demonstration of implementation, consider an agent which values health (\(v_1\)) and money (\(v_2\)). The agent is able to perceive two stimuli: its work (\(s_1\)) and vacation (\(s_2\)). Its judgement network \(J\) which, given its current internal and external states, has derivatives
\begin{align*}
    \begin{bmatrix}
        \frac{\partial v_1}{\partial s_1} & \frac{\partial v_1}{\partial s_2} \\
        \frac{\partial v_2}{\partial s_1} & \frac{\partial v_2}{\partial s_2}
    \end{bmatrix}
    &=
    \begin{bmatrix}
        0.3 & 0.9 \\
        0.8 & -0.2
    \end{bmatrix}
\end{align*}
To leverage the benefits of agent-based models (ABMs) and large language models (LLMs) \cite{grossmann_ai_2023}, these values are then input into the GPT 3.5 LLM. \cite{openai_gpt-35_2021} \\
\subsubsection{Input}
If Agent 1 is thinking about going on vacation, how will it feel?
\newline
\subsubsection{Output}
If Agent 1 is thinking about going on vacation, based on their beliefs, they would feel positively about the idea. Vacations are associated with a significant increase in valence for Agent 1 in terms of their health. While there may be a slight negative impact on their valence in terms of money, the overall sentiment would still be positive because of the significant positive impact on their health. Therefore, Agent 1 would likely feel excited, happy, and rejuvenated at the thought of going on vacation. \\
\subsection{Two-Agent Analysis}
A similar analysis can be performed with two agents. A second agent is considered which has identical values except that the first element in the matrix is defined as \(\frac{\partial v_1}{\partial s_1} = -0.5\). %
\subsubsection{Input}
Simulate a short, casual conversation between Agent 1 and Agent 2 about going to work. (Don't use numbers)
\subsubsection{Output}
\textbf{Agent 1:} Hey, Agent 2! Ready to dive into another workday?%
\\
\textbf{Agent 2:} Hey, Agent 1! Can't say I'm overly thrilled, but duty calls. How about you? Feeling motivated?%
\\
\textbf{Agent 1:} Haha, well, let's just say I'm trying to summon my inner motivation. But hey, we make a good team, right? We'll tackle it together!%
\\
\textbf{Agent 2:} Absolutely! Teamwork makes the dream work, they say. Maybe we can find some small joys amidst the tasks today. Any ideas?%
\\
\textbf{Agent 1:} Hmm, how about a little music to groove to while we work? A good playlist can turn even the most mundane tasks into a mini dance party.%
\\
\textbf{Agent 2:} I like the way you think, Agent 1! Music always helps lighten the mood. Count me in for the impromptu office dance-off.%
\\
\textbf{Agent 1:} Haha, that sounds like a plan! And you know what? Let's take a break together and grab a cup of coffee later. A little treat to keep us going.%
\\
\textbf{Agent 2:} Absolutely! Coffee is the fuel that keeps us moving forward. Let's savor that much-needed break and chat about something other than work for a change.%
\\
\textbf{Agent 1:} I couldn't agree more, Agent 2. Work doesn't have to be all serious business. Let's make the most of today, find those moments of joy, and make the workday a little brighter.%
\\
\textbf{Agent 2:} I'm with you on that, Agent 1. Let's embrace the challenges, find those pockets of fun, and support each other through the workday. We've got this!%
\\

\subsection{Analysis}
The one-agent example provides a means of analyzing an agent's emotions. The two-agent example provides a simple analysis of discussion of trade offs. This could be extended further to capture ideologies and cultural tendencies with larger agent populations. \cite{diallo_human_2019} Moreover, only one network was considered in this example, the other three networks in the model may provide more depth of analysis. The tone in which this conversation happened is also noteworthy. The tendency to work together and cooperate may potentially be an artifact of the values of the LLM itself.

\section{Ethical Considerations}
\label{appendix:ethics}
For the dynamic case, there exist sequences of stimuli, i.e. experiences, which can grow the diameter of the association horizon of a given stimulus. Beyond a certain degree of belief aggregation, other stimuli are sufficiently associated with it such that the agent's actions are largely a reaction to this stimulus. When considering a stimulus with a negative correlation to the other elements of self worth (\(\underline{\alpha}\)) and relative self efficacy (\(\underline{\beta}\)), traversal of its association horizon by a sufficient number internal stimuli often results in suicide. Conversely, there exist sequences of stimuli which can decrease the relative diameter of a stimulus association horizon or increase the diameter of others such that the relative effect of a given stimulus is lessened. These value-formation and value-dilution sequences can be used to treat patients or to efficiently train neural networks.

With the potential for malicious or exploitative sequences, machines which are constantly learning may require security patches which account for the potential injection of toxic correlations. For stability, community votes on the policies of safety-critical or judicial AIs, may help remove bias from networks purely trained on past examples.

In the dynamic case, generative adversarial techniques may be useful in expanding the range of stimuli with which an agent can cope. The therapeutic case may present an efficient method for the incremental treatment of trauma. Beyond this, the discovery of rectification circuit architecture may lead to the design of neuroware which is resilient to negatively-valenced stimuli.

Work remains for the optimal design of stimuli which, upon perception, produce actions and reactions which, when perceived, are detected as the original stimuli produced. Further study of this self-replicating behavior with this framework may clarify the nature of such eigenstimuli, i.e. habit triggers, and psychological viruses. The dynamics field may present avenues for the creation of treatments which mutate and evolve over time to keep pace with social viruses in a given population.

These possibilities may require great care to ensure ethical treatment of agents under consideration.

\section{Conclusion}

Emotions presented in this model depend on core value derivatives with respect to perceived stimuli in conjunction with stimulus associations and degree of perception.
In general, action on a system generates a stimulus.
\begin{equation}
\tilde{\underline{s}} = M(\tilde{\underline{a}})
\end{equation}
This external stimulus is then classified as a internal stimulus.
\begin{equation}\label{eq:stimulus-equation}
\underline{s} = C(\underline{\tilde{s}})
\end{equation}
This internal stimulus is then appraised by an agent.
\begin{equation}\label{eq:value-equation}
\underline{v} = (J \circ C)(\underline{\tilde{s}})
\end{equation}
This appraisal leads to action.
\begin{equation}
\underline{\tilde{a}} = (D \circ J \circ C)(\underline{\tilde{s}})
\end{equation}
The action then results in the generation of new stimuli.
Emotion can then be mapped onto a basis set characterized by three types of parameters. Valence results from the evaluation of a stimulus with respect to a given core value.
\begin{equation}
    \eta_{i,c} = \frac{\partial v_c}{\partial s_i}
\end{equation}
Perceived correlation accounts for the association of stimuli.
\begin{equation}
\label{eqn:correlation2}
\rho_{j,i} = \frac{\partial s_j}{\partial s_{i'}}
\end{equation}
Degree of perception determines effective arousal.
\begin{equation}
\Delta s_i = \langle s_{i} | (\underline{s} - \underline{s}_{ref})
\end{equation}

With a lower degree of detail, this framework approximates interaction of an agent with conceptual prototypes. With finer detail, this model can be granularized down to the level of neurons. Moreover, this formulation lends itself to the hypothesis that the process of mapping a stimulus from one layer to the next is automatic whereas the excitation of internal stimuli and modulation of weights and biases can be voluntary. However, the purpose of this work is not to make an argument for or against dual-process theory, as these concepts are discussed at length in other works. \cite{glockner_beyond_2010,kruglanski_persuasion_1999}

With the use of neural networks in this model, emotions of existing AIs can be directly computed without the definition and tuning of additional parameters. This may lead to considerations for ethical treatment of AI agents. The definition of emotions as proposed makes provision for high-association-radix emotions than those commonly considered, which provides additional flexibility in designing treatment plans for patients. Principally, the static case presented lays the foundation for emotion architecture dynamics and the observation and design of emotional trajectories in human and artificial agents.

\bibliography{00ProjectZimbardo}

\end{document}